\documentclass[10pt,twocolumn,letterpaper]{article}

\usepackage[accsupp]{axessibility}
\usepackage[pagenumbers]{wacv} 

\usepackage{graphicx}
\usepackage{amsmath}
\usepackage{amssymb}
\usepackage{booktabs}
\usepackage{makecell}
\usepackage{comment}
\usepackage{multirow}
\usepackage{color, colortbl}
\definecolor{LightCyan}{rgb}{0.88,1,1}

\usepackage[capitalize]{cleveref}
\crefname{section}{Sec.}{Secs.}
\Crefname{section}{Section}{Sections}
\Crefname{table}{Table}{Tables}
\crefname{table}{Tab.}{Tabs.}

\usepackage{xr}
\makeatletter

\begin{document}

\title{HaGRID -- HAnd Gesture Recognition Image Dataset}

\author{
Kapitanov Alexander$^*$ \\
{\tt\small kapitanovalexander@gmail.com}
\and
Kvanchiani Karina$^*$ \\
{\tt\small karinakvanciani@gmail.com}
\and
Nagaev Alexander$^*$ \\
{\tt\small sashanagaev1111@gmail.com}
\and
Kraynov Roman$^*$ \\
{\tt\small ranakraynov@gmail.com}
\and
Makhliarchuk Andrei$^*$ \\
{\tt\small helloworld106@gmail.com}
\\
\\
\hspace{-12em}SaluteDevices, Russia
}

\maketitle
\def\thefootnote{*}\footnotetext{These authors contributed equally to this work.}
\begin{abstract}
This paper introduces an enormous dataset, HaGRID (HAnd Gesture Recognition Image Dataset), to build a hand gesture recognition (HGR) system concentrating on interaction with devices to manage them. That is why all 18 chosen gestures are endowed with the semiotic function and can be interpreted as a specific action. Although the gestures are static, they were picked up, especially for the ability to design several dynamic gestures. It allows the trained model to recognize not only static gestures such as ``like" and ``stop" but also ``swipes" and ``drag and drop" dynamic gestures. The HaGRID contains 554,800 images and bounding box annotations with gesture labels to solve hand detection and gesture classification tasks. The low variability in context and subjects of other datasets was the reason for creating the dataset without such limitations. Utilizing crowdsourcing platforms allowed us to collect samples recorded by 37,583 subjects in at least as many scenes with subject-to-camera distances from 0.5 to 4 meters in various natural light conditions. The influence of the diversity characteristics was assessed in ablation study experiments. Also, we demonstrate the HaGRID ability to be used for pretraining models in HGR tasks. The HaGRID and pre-trained models are publicly \def\thefootnote{\arabic{footnote}}available\footnote{\url{https://github.com/hukenovs/hagrid}}\footnote{\url{https://gitlab.aicloud.sbercloud.ru/rndcv/hagrid}}.
\end{abstract}

\section{Introduction}
\label{Introduction}

Using gestures in human communication plays a vital role~\cite{usage}: they can emotionally reinforce statements or entirely replace them. Moreover, Hand Gesture Recognition (HGR) can be a part of human-computer interaction to determine the gesture a person shows to the camera and perform an action corresponding to it. Since people universally use gestures in real life, building HGR systems can improve user experience and accelerate processes in such domains as the automotive sector~\cite{auto1},~\cite{auto2}, home automation systems~\cite{home}, multimedia applications, a wide variety of video/streaming platforms (Zoom, Skype, Discord, Jazz, etc.), and others~\cite{others1},~\cite{others2}. Besides, such a system can be a part of a virtual assistant or service for active sign language users – hearing and speech-impaired~\cite{defects1},~\cite{defects2}.

\begin{figure}[t]
  \centering
  \includegraphics[width=1.0\linewidth]{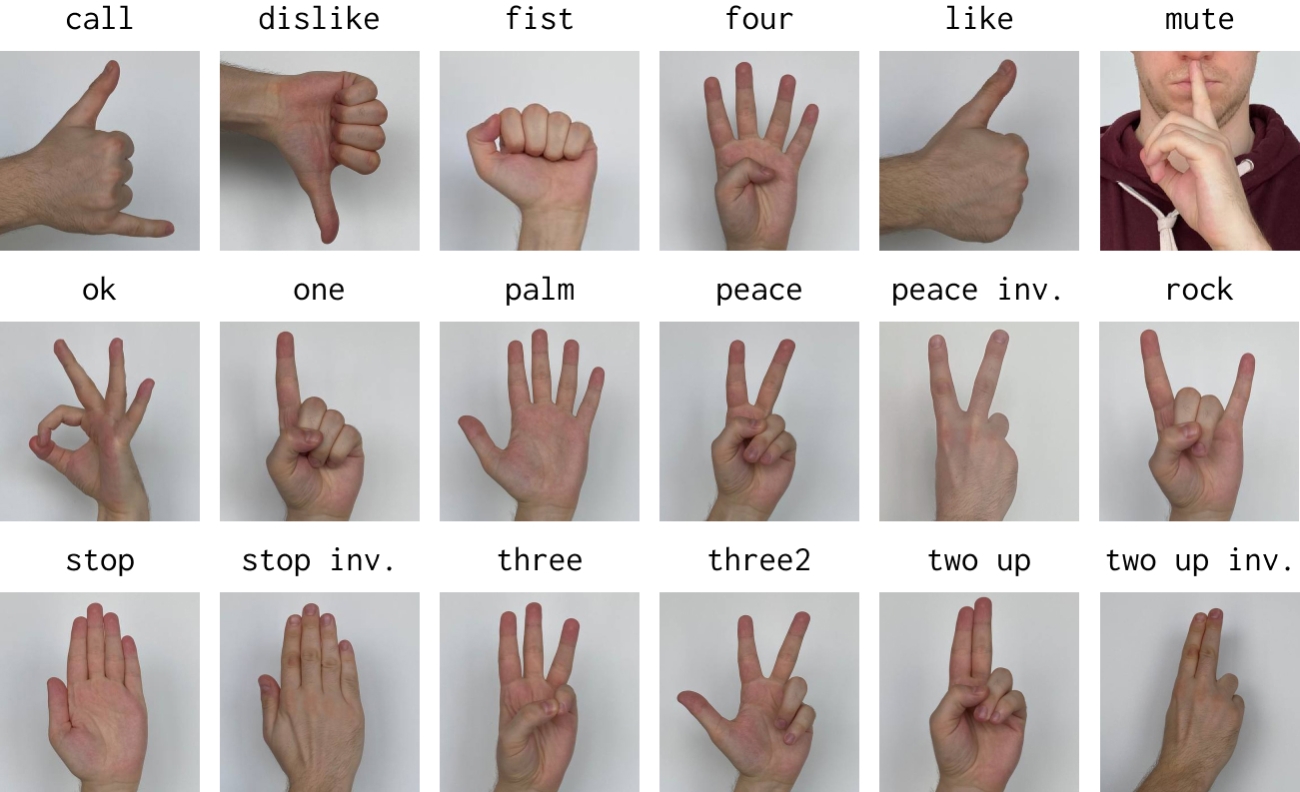}
  \caption{The 18 gesture classes included in HaGRID (``inv." is the abbreviation of ``inverted").}
  \label{fig: gestures}
\end{figure}

\begin{table*}
  \centering
  \scalebox{0.8} {
  \begin{tabular}{@{}lccccccc@{}}
    \toprule
    Dataset & Samples & Classes & Subjects & Scenes & Resolution & Annotations & Annotation Method\\
    \midrule
    LaRED, 2014~\cite{lared} & 243,000 & 81 & 10 & 10 & 640 × 480 & masks & automatically \\
    OUHANDS, 2016~\cite{ouhands} & 3,000 & 10 & 23 & various & 640 × 480 & masks, boxes & automatically \\
    HANDS, 2021~\cite{hands} & 12,000 & 29 & 5 & 5 & 960 × 540 & boxes & --\\
    SHAPE, 2022~\cite{shape} & 33,471 & 32 & 20 & various & 4128 × 3096 & masks, boxes & manually \\
    \rowcolor{LightCyan}
    HaGRID, 2023 & \textbf{554,800} & 18 + 1 & \textbf{37,583} & 
    \textbf{$\geqslant$ 37,583} & 1920 × 1080 & boxes & manually \\
    \bottomrule
  \end{tabular}}
  \newline
  \caption{The main parameters of the mentioned hand static gesture datasets. + 1 in the third column means the dataset contains an extra class ``no gesture". The number of scenes in the last row cannot exceed the number of subjects. Note that the HaGRID consists of at least 90\% FullHD images. The information about the annotation method for the HANDS dataset was not found.}
  \label{tabl:datasets1}
\end{table*}

The primary objective of our study was to build the HGR system for the following implementation in home automation devices with virtual \def\thefootnote{\arabic{footnote}}assistants\footnote{\url{https://sberdevices.ru/sberportal/}}\footnote{\url{https://sberdevices.ru/sberboxtop/}} and the video conferencing service \def\thefootnote{\arabic{footnote}}Jazz\footnote{\url{https://developers.sber.ru/portal/products/jazz-by-sber}}. Primarily, the set of gestures must be intuitive~\cite{intuitive} and straightforward, so that system users can remember them for comfort interaction. Also, the HGR system should be designed with gestures suitable for controlling it; frequently, these are gestures with semiotic and ergotic functions~\cite{roles}. Semiotic gestures aim at sharing information, in our case, between humans and computers, to receive a system response and can be static or dynamic. In comparison, the ergotic gestures are manipulated with objects (e.g., swipe or drag and drop something) and can only be dynamic. Given the above and the necessity of a real-time system in our domains that uses lightweight models, datasets containing images with static gestures are more suitable. Besides, related applications require that the HGR system exclusively make decisions based on gestures, i.e., be robust to the amount of context in images, background, subjects, and lighting conditions. 

This paper presents the HaGRID dataset to design the above HGR system for home automation devices and services because existing datasets' characteristics are insufficient (Section~\ref{Related Work}). The proposed dataset contains more than half a million images divided into 18 classes of not-language-oriented gesture signs (Fig.~\ref{fig: gestures}). Such gestures are chosen to design a device control system and serve one semiotic functional role. Section~\ref{Dynamic} of this paper presents a methodology for designing dynamic ergotic gestures by combining a set of static semiotic ones. A small lexicon of the most comfortably designed actions in the dataset is conceived to reduce HGR system complexity and avoid unnecessary cognitive load on the device user. We also added an extra class with samples of natural hand movements and called it ``no gesture” to avoid false positive triggering. Remarkably, our dataset consists of many images per class, all with considerable context, which differs in background, lighting, scene, and subjects. This heterogeneity is achieved using two crowdsourcing platforms, namely, \def\thefootnote{\arabic{footnote}}Yandex.Toloka\footnote{\url{https://toloka.yandex.ru}} and \def\thefootnote{\arabic{footnote}}ABC Elementary\footnote{\url{https://elementary.activebc.ru}}. The dataset creation pipeline is also provided in this paper as a contribution.

The HaGRID was annotated by bounding boxes to (1) design dynamic manipulative gestures, (2) error-free recognition at long distances, and in cases where there are several people in the frame, (3) simplify full-frame hand gesture classification task by reducing it to cropped hand image classification. Also, we paid attention to other gesture user experiences, i.e., the beat dancer \def\thefootnote{\arabic{footnote}}app\footnote{\url{https://apps.sber.ru/salute-apps/}} implemented in our devices -- this requires recognizing both hands in the frame, which is impossible without box markup. Besides, bounding box annotations are more stable than keypoint annotations under challenging conditions such as extremal lighting and large subject-to-camera distance. 

In Section~\ref{Ablation Study} of this paper, we provided the set of dataset ablation experiments to explore the degree of dataset characteristics' influence on the result of solving the HGR as classification and detection problems. Besides, we conducted experiments to demonstrate that the HaGRID can be a sufficient dataset for pretraining HGR models with the following finetuning.

\section{Related Work}
\label{Related Work}
\subsection{Hand Gesture Datasets}
There are at least 50 hand gesture recognition datasets. Their gesture baskets can be divided into 3 main groups of style~\cite{styles}: sign language~\cite{asl}~\cite{chalearn1}, semaphores~\cite{cambridge}~\cite{chalearn2}~\cite{lared}~\cite{hands}~\cite{ouhands}~\cite{shape}, and manipulation gestures~\cite{ego}~\cite{nvidia}~\cite{chalearn2}. The first group's datasets propose complex dynamic gestures, which are more applicable for their original purpose and redundantly for our goals demanding straightforward actions. The last two groups find applications in home automation systems and human-computer interaction and perform semiotic and ergotic roles accordingly. As we aimed to build an HGR system with a predominantly semiotic role by adding manipulative gestures solely using heuristics, only datasets with static gestures are reviewed in this section. 

Since the HGR system users presumably will show gestures at a distance from the device, the trained model needs to capture the whole context and search for a person's hand in it. However, some datasets with static gestures are intended for person-independent systems and contain samples of no human body with only hand parts, i.e., cropped hand images~\cite{indian}~\cite{arasl}, which is why they are unsuitable for us. Static gesture datasets are frequently annotated with the following markup types or their combinations: class labels, bounding boxes, keypoints, and segmentation masks. Only class annotations are insufficient for us due to the need for error-free work on the multiple-hand frames. Segmentation masks are redundant and unsuitable for this task as they are not intended to classify objects so similar as hand gestures well, whereas keypoints are impossible to use as they stick together over long distances. To our knowledge, there are only 4 datasets for static gesture recognition with context and appropriate annotations, including HANDS~\cite{hands}, SHAPES~\cite{shape}, OUHANDS~\cite{ouhands}, and LaRED~\cite{lared}. 

They differ by the number of samples, image resolution, the number of classes, the presence of negative samples, the homogeneity of scenes, and the distance between the camera and each subject. The SHAPE and the OUHANDS are marked by bounding boxes and segmentation masks; the LaRED are marked only with masks, and the HANDS -- only with bounding boxes. This paper discusses datasets for solving only hand gesture classification and hand detection problems without segmentation.

The mentioned datasets are not appropriate for constructing our HGR system due to the insufficiency of heterogeneity in such characteristics as scenes and subjects, which negatively affects the heterogeneity in lighting conditions and subject-to-camera distances. In the ablation study (Section~\ref{Ablation Study}), the experiments proving the necessity of such characteristics for neural network generalization are provided. 
Besides, there are other disadvantages to each of them:
\begin{small}
\begin{itemize}
\item The LaRED dataset~\cite{lared} is divided into 27 main classes of gestures and 54 additional classes created by rotating the primary gestures about two axes. It was collected by a short-range depth camera, which implies a small context amount in the images; therefore, the trained model is mistaken at a significant subject-to-camera distance. Besides, each subject performed 300 images per class with only slight hand movements, making these images almost identical. Unfortunately, we could not obtain this dataset due to an outdated link.
\item The OUHANDS dataset~\cite{ouhands} was created to streamline the testing process for Human-Computer Interaction (HCI) tools with 10 unique categories, each containing 300 images. However, training a robust model for our particular task may not be achievable with this dataset. Different recording conditions are only able to improve the situation partially. Besides, most of the subject's hands are close to the camera.
\item The HANDS~\cite{hands} is a dataset for human-robot interaction consisting of 29 suitable for this application gestures, which are simple and easy to use. However, most of them differ little, complicating the use of the HGR system. The authors consider lighting conditions; nevertheless, it cannot be sufficient for dataset variability with only 5 backgrounds.
\item The gesture classes of the SHAPE~\cite{shape} dataset are chosen with a focus on the meaning; however, some gestures are specific in terms of culture, which limits their usefulness to people from other countries. The authors varied external factors during the recording samples and adapted them to the specific domain of mobile-related development by changing sides of taking photos. Despite this, SHAPE is not diverse in subjects, and some gestures are not intuitive for device users. Note that the SHAPE is not publicly available, and we could not get them upon request from the authors.
\end{itemize}
\end{small}

The above limitations push us to create a new HGR dataset with no such weaknesses. From Table~\ref{tabl:datasets1}, one can see that the proposed dataset is the largest in sample number and has the highest diversity scores across subjects and scenes, which helps avoid overfitting. 

\begin{figure*}[t]
  \centering
  \includegraphics[width=1.0\linewidth]{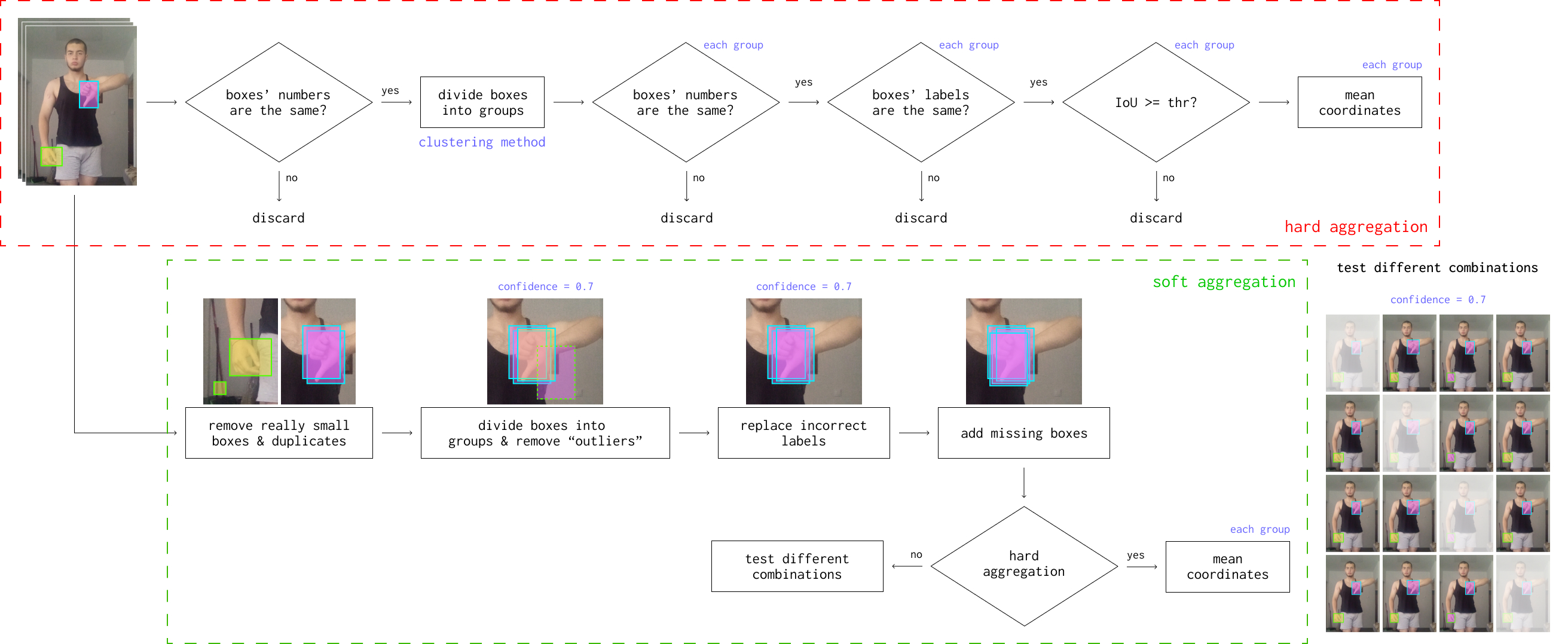}
  \caption{Bounding box aggregation pipeline. For hard aggregation, consistency checks are applied for all markups before averaging. If it fails, soft aggregation prepares for successful hard aggregation.}
  \label{fig: agg}
\end{figure*}

\subsection{Dataset Creation Pipeline}
Since the dataset creation pipeline is one of our contributions, we reviewed existing pipelines of collecting and annotating HGR datasets. The following creation methods differ in the choice of subjects and their quantity, depending on their number, the method of recording and transferring samples, and data diversification and annotation methods.

Most of the reviewed datasets were created by manual recording, which prevented them from being more heterogeneous and containing enough samples. To improve the situation slightly, the dataset's authors attempted to diversify the data. As an illustration, a variety of the HANDS dataset was achieved through (1) 5 different backgrounds, (2) the presence of cluttered and uniform backgrounds while the subject is standing still or moving, and (3) artificial and natural lights. The SHAPE was diversified by varying subject-to-camera distances, backgrounds, and clothing. LaRED dataset's authors optimized data recording using their software tool. However, the requirement for subjects to record 100 frames for each gesture three times entails low data variability. The HANDS and the SHAPE operate one of the tricks in creating data in such limited conditions performing the same gesture with both hands to optimize the collecting and annotating of images and increasing the number of classes by 2 times (right-hand and left-hand gestures are 2 different classes). Regarding the labeling process, as we know, only the SHAPE dataset was annotated manually. Segmentation masks in the OUHANDS and the LaRED were generated using depth images, which entails the roughness of this markup.

The most significant limitations of existing static gesture datasets are homogeneous context, restricted subject amount, and insufficient number of samples to train robust HGR models. It is affected by the creation of a datasets pipeline founded on manual recording in the controlled lab environment~\cite{crowd}. We utilize crowdsourcing platforms to overcome these restrictions and build close-to-real distribution variant data. The authors~\cite{crowd} underlined that this choice of data creation method can enhance recognition performance. 

\section{HaGRID Dataset}
\label{HaGRID Dataset}
The need for a combination of such characteristics as (1) high-resolution images, (2) heterogeneity across the image scene, subjects, their age and gender, lighting, subject-to-camera distance, (3) a sufficient number of samples, and (4) static and functional gestures became the motivation for creating the HaGRID and involving crowdsourcing platforms for it. The dataset comprises more than half a million predominantly FullHD RGB images, with the most suitable for our domain 18 gestures and a ``no gesture" class. The dataset was recorded with 37,583 subjects and at least an equal number of unique scenes, displaying heterogeneity in other characteristics. In addition to class division, the HaGRID is annotated by bounding boxes for the hand detection problem: each image has $n$ corresponding bounding boxes for $n$ hands in a frame, where $n \in [1, 2]$.

\subsection{Dataset Creating Pipeline}
The dataset creation pipeline is described step by step to showcase how heterogeneity is achieved and to provide details on the dataset's content and quality. The dataset was created in 4 stages: (1) the image collection stage called mining; (2) the validation stage, where mining rules and some conditions are checked; (3) the filtration of inappropriate images; and (4) the annotation stage for markup bounding boxes. The classification stage is built into the mining, validation, and annotation pipelines by splitting pools for each gesture class. We use two Russian crowdsourcing platforms: Yandex.Toloka (1, 2, and 4 steps) and ABC Elementary (3 and 4 steps) to complete these stages. Note that all crowd workers are aware of the prohibition on the transfer of personal data to third parties and the presence of dubious content. Using two platforms at the annotation stage allows us to increase the final annotation confidence due to the two different annotator domains' involvement. The details of each of the steps are as follows:

\begin{figure*}[t]
  \centering
  \includegraphics[width=1.0\linewidth]{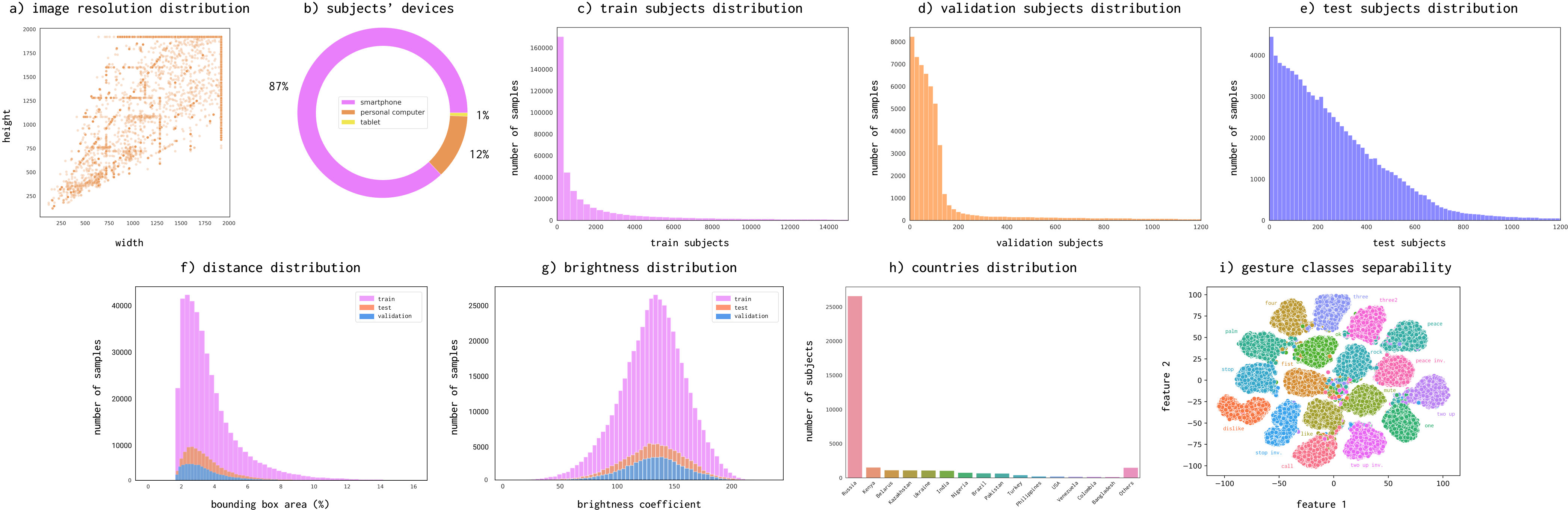}
  \caption{Image resolution, brightness, subject-to-camera distance, subjects, and class separability analysis. a) image resolution distribution: samples overlap with equal transparency and density reveals quantity, the minimum dimension of 90\% images is 1,080; b) subjects' devices: only smartphones, personal computers, and tablets were used while recording; c), d), e) image distribution by subjects in train, validation, and test sets, respectively; f) subject-to-distance distribution: distance was computed as bounding box area relative to the whole image (the boxes occupy up to 16\% of the image); g) brightness distribution: images were converted to grayscale, and average pixel brightness was received; h) subjects' countries distribution; i) t-SNE plot by ResNet-18 features.}
  \label{fig: chars}
\end{figure*}

\textbf{1. Mining.} The crowd workers' task was to take a photo of themselves with the particular gesture indicated in the task description. We define the following criteria: (1) the annotator must be at a distance of 0.5 -- 4 meters from the camera, and (2) the hand with gesture must be entirely in the frame. Sometimes, we altered lighting conditions from low light to a bright light source to make the neural network resilient to extreme cases. Periodically, we changed countries in mining tasks on the crowdsourcing platform, covering more ethnic groups due to their correlation with countries. All received images were also checked for duplicates using image hash comparison~\cite{hash}. The mining tasks were accompanied by instructions with a warning about the further publication of the crowd workers' photos.

\textbf{2. Validation.} 
We implemented the validation stage to achieve high-confidence images by removing those where the conditions for the mining stage were not fulfilled. The validation stage aims to favor correctly executed images at the mining stage, i.e., classify them with classes ``correct" and ``incorrect"; only ``correct" images get into the dataset. For the high-quality validation, we operated tricks such as access to the main tasks after training and exams and using control tasks to prevent crowd workers from cheating. For each image at this stage, we set in the system the dynamic overlap of 3 to 5 performers, i.e., each assignment was completed by at least three crowd workers. Based on the majority rule, some photos were rejected, while the rest have been passed to the filtration stage. After the validation stage, about 70\% of images remained for each gesture.

\textbf{3. Filtration.}
For ethical reasons, images of children, people without clothes, and images with inscriptions were removed from the HaGRID at this stage. We use a solid rule for the filtration stage -- 5 workers should filter each image. For an answer to be accepted, it must receive at least four positive votes from workers. Similar to the validation stage, annotators pass a thorough exam, training, and control tasks at the filtration stage. More than 85\% images passed the filtration stage.

\textbf{4. Annotation.} At the annotation stage, after passing the exam, crowd workers should draw a bounding box around the gesture on each image and another one around the hand without the gesture if it is entirely in the frame with specific labels (``gesture" or ``no gesture"). Annotation overlap is placed dynamically from 3 to 5 in each crowdsourcing platform. All markups, ranging from 6 to 10, are collected from two platforms and aggregated by one of the two schemes -- hard and soft aggregation algorithms (see Fig.~\ref{fig: agg}). About 5\% of images are not aggregated after the maximum overlap is not included in the dataset.

\begin{table*}[]
\centering
\scalebox{0.9}{
\begin{tabular}{lccccc}
\hline
\multirow{2}{*}{Model} & 
\multirow{2}{*}{Model size (MB)} & 
\multirow{2}{*}{Parameters (M)} & 
\multirow{2}{*}{Inference time (ms)} & 
\multicolumn{2}{c}{Metrics} \\ 
\cline{5-6} &  &  &  & F1-score & mAP \\
 \hline
ResNet-18 & 89.6 & 11.2 & 49.25 & 97.5 & - \\
ResNet-152 & 466.5 & 58.3 & 292.6 & 95.5 & - \\
ResNeXt-50 & 184.6 & 23.2 & 135.6 & \textbf{98.3} & - \\
ResNeXt-101 & 696.4 & 87 & 397.2 & 97.5 & - \\
MobileNetV3 small & 12.5 & 1.6 & 10.6 & 86.4 & - \\
MobileNetV3 large & 34 & 4.3 & 33.4 & 91.9 & - \\
VitB16 & 686.6 & 85.9 & 325.5 & 91.1 & - \\
\hline
RetinaNet ResNet-50 & 294.2 & 38.2 & 235 & - & \textbf{79.1} \\
SSDLite MobileNetV3 small & 9.4 & 1.9 & 30.7 & - & 57.7 \\
SSDLite MobileNetV3 large & 20 & 3.4 & 52.5 & - & 71.6 \\
YoloV7 tiny & 49 & 6 & 14.4 & - & 71.6 \\
\hline
\end{tabular}}
\newline
\caption{Models' training results on the HaGRID. F1-score and mAP (mean Average Precision) were chosen as the classification and detection metrics, respectively. Intel(R) Xeon(R) Platinum 8168 CPU @ 2.70GHz is used for computing inference time.}
\label{tabl:results}
\end{table*}

\subsection{Dataset Characteristics}

\textbf{Size and Quality.} HaGRID size is approximately 770 GB -- it includes more than 550 thousand images divided into 18 most intuitive classes of gestures: ``call", ``dislike", ``fist", ``four", ``like", ``mute", ``ok", ``one", ``palm", ``peace", ``peace inverted", ``rock", ``stop", ``stop inverted", ``three", ``three2", ``two up", ``two up inverted" (shown in Fig.~\ref{fig: gestures}). Since the HaGRID was designed to control devices or device apps, gestures are endowed to raise specific associations due to their meaning (see Table~\ref{tabl:rules} in the supplementary material). Such gestures allow us to solve particular problems, such as like/dislike something by relevant signs, play/stop the recording by ``peace" and ``stop", turning on/off the sound by ``peace" and ``mute", controlling the adjustable scale (e.g., volume scale) by ``one", ``peace", ``three", ``four", ``palm" and their combinations, etc. In addition, the user can combine some static gestures to create a new dynamic gesture not included in the dataset (Section~\ref{Dynamic}). Each gesture class contains more than 30,000 high-resolution RGB images (Fig.~\ref{fig: chars}a). 

\textbf{Content.} The HaGRID was recorded by 37,583 unique faces in at least as many unique scenes. The subjects' ages vary from 18 to 65 years old and are gender balanced. The subjects are primarily from Russia and, to a lesser degree, 115 other countries; this distribution is proposed in Fig.~\ref{fig: chars}h. We considered the scene specifics of such applications as home automation and video conferencing services, and we preferred mainly indoor context with considerable variation in lighting, including artificial and natural light. Besides, the dataset includes images taken in extreme conditions, such as facing and standing back to a window (see Fig.~\ref{fig: chars}g). Also, the subjects demonstrated gestures at different distances from the camera (Fig.~\ref{fig: chars}f) of the smartphone, personal computer, or tablet (Fig.~\ref{fig: chars}b). All images contain context information that is significant for our applications (see Fig.~\ref{fig: samples} in the supplementary material). The mean and standard deviation of HaGRID images' pixel values are equal [0.54, 0.499, 0.473] and [0.231, 0.232, 0.229], respectively. 

\textbf{Annotations.} The HaGRID was annotated by bounding boxes, the optimal annotation type for our applications. Such a choice allows us to train lightweight hand gesture detectors or recognize swipes and other dynamic gestures for interaction with objects on the device's screen. Each image was annotated by at least one box for a hand with a gesture. If the second hand is in the frame -- the bounding box is provided for it with the extra class ``no gesture". Although the ``no gesture" hands are predominantly passive and thus similar to each other, it's sufficient to eliminate primitive false positive errors (see the demo in the repository). We plan to diversify the extra class by adding samples with natural hand movements similar to target gestures in future dataset versions. Only 108,056 images contain a bounding box with an extra class. Bounding box annotations are proposed in COCO~\cite{coco} format with normalized relative coordinates.

\textbf{Splitting.} The dataset was split into training (74\%), validation (10\%) and testing (16\%) sets by subject. The subjects in training, validation, and testing sets equal 33,966, 1,908, and 1,709, respectively. Figure~\ref{fig: chars}c-e indicates that the test and validation sets were purposely designed to be more heterogeneous in subjects than the train set for the most representative results. Each set preserved the original distributions of brightness, subject-to-camera distance, age, and gender due to their random sampling.

In addition, the anonymized user ID hash is proposed in the annotation file, which allows the researchers to split the HaGRID themselves. Since the dataset size is large, we designed a small version (100 samples per class) of the HaGRID with annotations for preview at the link for its user comfort. For the same reason, the downscaled version (where the maximum image dimension is 512) with a 26 GB size is available. Dataset users can take advantage of automatically generated keypoint annotations by MediaPipe~\cite{mediapipe} to train hand estimation models. Besides, keypoint annotations can be used to pre-train the model on the HaGRID and finetune on the other hand gesture classes.

\section{Base Expirements}
To assess the capabilities of the dataset, we evaluated 11 popular architectures of heterogeneous size and number of parameters for the two HGR tasks: hand detection and hand gesture classification. We chose SSDLite with MobileNetV3 small and large backbones~\cite{mobilenets2}, RetinaNet with ResNet50 backbone~\cite{retina} and YoloV7 tiny~\cite{wang2022yolov7} as detectors, and set of 7 architectures consisting of ResNet-18, ResNet-152~\cite{resnets}, ResNeXt-50, ResNeXt-101~\cite{resnext}, ViTB16, MobileNetV3 small and MobileNetV3 large~\cite{mobilenets3} as classifiers.

\subsection{Experiment Setup}
Due to the large dataset size, each model, except pre-trained on ImageNet ViTB16, was trained from scratch on full-frame images. The metrics below were calculated on the testing set containing 90,000 images. We downsampled images on the maximum side to a size of 224 and padded the minimum side to 224. The models were trained on a single Tesla V100 with 32GB with a batch size of 128 to convergence -- an early stop was triggered if the metric did not increase by at least 0.01 after 10 epochs. The training set-up for the models is summarized in Table~\ref{tabl:params} in supplementary material. Note that the ``no gesture" class is utilized only in the detection task, while full-frame classification is based on 18 main classes due to each image containing one of the target gestures.

\subsection{Results}
Table \ref{tabl:results} presents the evaluation results of the selected model architectures for solving gesture detection and classification problems. Such high performance demonstrated the dataset's ability to train models without added complexities in the training stage. The demo of our gesture recognition system solving classification and detection tasks is available in our repository. It highlights the practical applications of training models on image datasets, such as real-time and video stream analysis, for product development.

\subsection{Dynamic Gesture Recognition}
\label{Dynamic}
\begin{figure}
  \centering
  \includegraphics[width=1.0\linewidth]{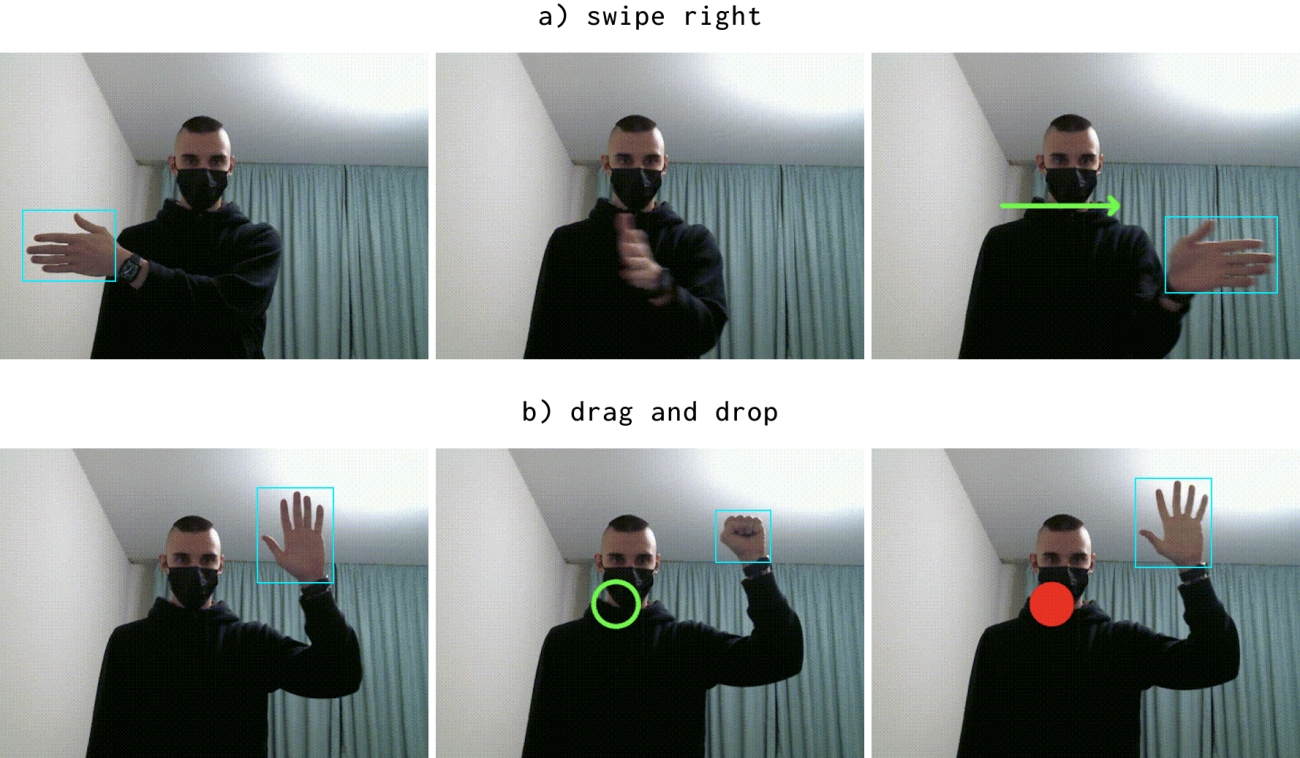}
  \caption{The screenshots from the dynamic gesture recognition demo: a) ``swipe right" gesture recognition occurred by detecting serial pair of left-rotated ``stop inverted" and right-rotated ``stop"; b) ``drag and drop" -- by detecting the subsequence: ``palm", ``fist" and ``palm".}
  \label{fig: dynamic}
\end{figure}

The observance of specific rules is applied to build a dynamic gesture recognizer using the dataset with only static gestures. The essence of this approach is to divide the dynamic gesture into two components: the initial and final gestures. For example, the dynamic gesture ``swipe right" consists of a left-rotated and a right-rotated gesture ``stop" as a start and as an end, respectively, while the gesture ``drag and drop" can be shown by ``fist" and ``palm" as a start-end pair (see Figure~\ref{fig: dynamic}).

We developed the gesture prediction queues to implement dynamic gestures as an empty list of a certain depth, filled with events on each frame. Queues verify the correctness of the execution of a dynamic gesture. The queue is replenished with found bounding boxes by hand detector and corresponding classes of gestures from the classifier. The recognition depends on the sequence of actions in the queue, time constraints between start and end gestures, and positional location of start and end gestures. After identifying a dynamic gesture, the queue is reset, and the process continues with the definition of static gestures. 

Since we need to detect both hand gestures and intermediate states of the hand and to recognize rotated gestures, the YoloV7 tiny detector and LeNet~\cite{lecun} as the lightweight classifier were utilized for the demo separately.

\section{Ablation Study for HaGRID}
\label{Ablation Study}

An ablation study was conducted to assess the main heterogeneity characteristics' impact individually. We tested the necessity for large amounts of data, diversity in brightness, subject-to-camera distances, and number of subjects by changing these characteristics and freezing the rest. In the ablation study, we utilized ResNet-18, ViTB16, and MobileNetV3 (small and large versions) for the classification task and SSDLite with both small and large MobileNetV3 and RetinaNet with ResNet50 for detection. Several training data modifications were sampled for each of the described characteristics to find the best one for all models. Validation and test sets were unchanged in all experiments.

In addition to checking the influence of the characteristics on the HaGRID test, we also decided to assess it on other data -- on the OUHANDS. As the HaGRID and the OUHANDS datasets do not intersect in gesture classes, we finetuned all models learned by different training data modifications on the OUHANDS and tested on its test set. The results also show the HaGRID ability to be the acceptable dataset for pretraining models for the static HGR task. 

\begin{figure*}
  \centering
  \includegraphics[width=1.0\linewidth]{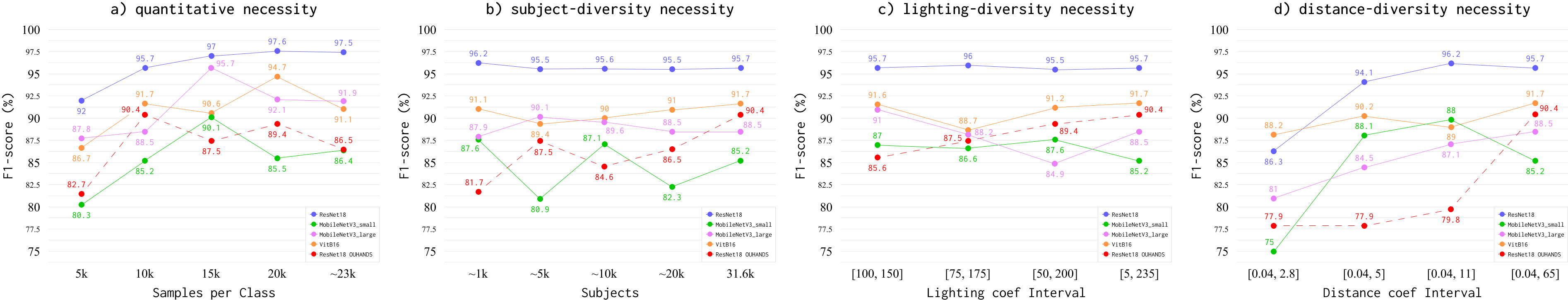}
  \caption{The impact visualization of such dataset characteristics as a) sample amount, diversity in b) subjects, c) lighting, and d) subject-to-camera distance to train accurate and resilient classifiers. Solid lines correspond to models trained and tested on the HaGRID dataset, whereas the dotted line is the model pretrained on the HaGRID, finetuned on the OUHANDS, and tested on its test set. The F1-score of the trained from scratch on the OUHANDS ResNet-18 is 60.6.}
  \label{fig: classification_metrics}
\end{figure*}

\begin{figure*}
  \centering
  \includegraphics[width=1.0\linewidth]{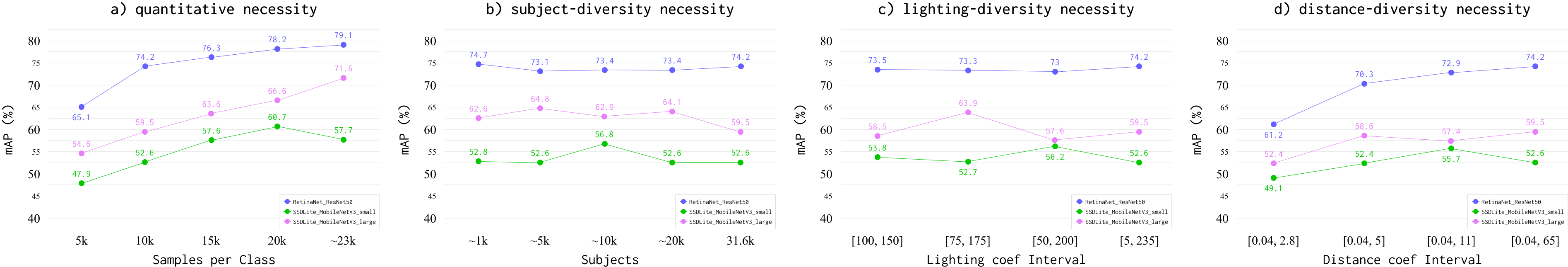}
  \caption{Similar to the graph above, the detectors were trained on data of various heterogeneity and quantity to assess the impact of dataset characteristics.}
  \label{fig: detection_metrics}
\end{figure*}

\subsection{Quantitative Necessity}
To assess the influence of the data amount, we trained 5 models per architecture with different sample numbers per class from 5,000 to all samples in steps of 5,000. The deterministic slice was used for a train set expansion, i.e., images in the $n[i]$ set are included in the $n[i + 1]$ set. The other heterogeneity characteristics retain their uniform distribution due to the premixing of data, which limits their influence and provides the interpreted results.

\textbf{Quantitative Necessity Results.} The quantitative necessity results for classifiers and detectors (see Fig.~\ref{fig: classification_metrics}a and Fig.~\ref{fig: detection_metrics}a) demonstrated an upward trend as the training set increases. On average, the enhancement increases rapidly at the beginning and less significantly towards the end. While approximately 23,000 samples per class are redundant for classifiers, then for detectors, they are essential and justified to achieve the best performance.

\subsection{Subject-Diversity Necessity}
The significance of the subject's quantity is also evaluated by varying the number of unique individuals in the training set. The set amount is fixed to 10,000 images per class for all diversity experiments; this number allows us to sample data with different heterogeneity, which is enough for high performance. The other 2 characteristics -- brightness and subject-to-camera distance distributions -- are also unchanged. We utilized a sampling algorithm for each class to vary the subject's quantity inside 10,000 images. This algorithm sorts the list by the number of images from a unique subject and moves toward the middle from the left and right at different speeds (depending on the required subject's quantity).

\textbf{Subject-Diversity Necessity Results.} Despite that, the trend is practically unchanged on the HaGRID test in classification and detection tasks (shown in Fig.~\ref{fig: classification_metrics}b and Fig.~\ref{fig: detection_metrics}b), the number of subjects has a positive effect on additional training on OUHANDS data.

\subsection{Lighting-Divercity Necessity}
Similar to the subjects' experiments, we varied lighting diversity inside 10,000 images. Four brightness coefficient windows were chosen from homogeneous lighting to heterogeneous: [100, 150], [75, 175], [50, 200], [5, 235]. The enormous amount of data allows us to maintain distributions of the rest of the features: subjects and subject-to-camera distance.

\textbf{Lighting-Diversity Necessity Results.} Figure~\ref{fig: classification_metrics}c and Figure~\ref{fig: detection_metrics}c show that lighting diversity is not a significant feature in the context of testing on the same dataset. However, finetuning on the OUHANDS dataset is most effective with more significant brightness heterogeneity.

\subsection{Distance-Diversity Necessity}
As with the lighting-diversity experiments, windowed sampling was applied to perform the ablation distances-diversity experiment. To vary the heterogeneity of the subject-to-camera distance, we selected windows with a static basis close to zero: [0.04, 2.8], [0.04, 5], [0.04, 11], [0.04, 65]. The distance coefficients were calculated as the ratio of the area of the bounding box to the area of the image:
\[distance = 100 \ast W \ast H,\]
where \(W, H\) are the width and the height of the gesture bounding box, respectively. Since bounding box annotations in the HaGRID are relativity, the division by the image area is omitted, and for perceiving convenience, we have multiplied the result by a constant equal to 100. As in other experiments, the training set contains 10,000 per class, and the distributions of the rest of the heterogeneous characteristics are saved.

\textbf{Distance-Diversity Necessity Results.} The classifiers' and detectors' performance depends on subject-to-camera distance diversity both for the HaGRID test and for the OUHANDS finetuning (see Fig.~\ref{fig: classification_metrics}d and Fig.~\ref{fig: detection_metrics}d).

\section{Conclusion}

In this paper, we introduce the HAnd Gesture Recognition Dataset called HaGRID, one of the largest and most diverse in subjects and context HGR datasets. It is mainly intended to be used in system control devices, but the potential for its application is quite vast. Heterogeneity in such characteristics as subjects, subject-to-camera distances, scenes, and lighting conditions positively influence the training of a resilient model. We also show the ability of selected classes of gestures to construct dynamic gestures and provide its recognition demo. Our following work with the HaGRID consists of increasing the gesture classes, samples with natural behaviors of users’ hands similar to the target gestures, and samples with different subjects' translations and rotations. The whole dataset, its downsampled version, the trial version with 100 images per class, pre-trained models, and the dynamic gesture recognition demo are publicly available in the \def\thefootnote{\arabic{footnote}}repository\footnote{\url{https://github.com/hukenovs/hagrid}}. 

{\small
\bibliographystyle{ieee_fullname}
\bibliography{hagrid}
}

\clearpage
\onecolumn
\appendix
\section*{Supplementary materials}

\centering
\begin{table}[htp]
\begin{center}
\scalebox{0.95}{
\begin{tabular}{|p{1.6in}|p{2.3in}|p{0.4in}|p{2.3in}|}
\hline
Gesture & Applications & Gesture & Applications\\
\hline
one, peace, three, three2, four, palm & numeric value input, e.g. volume scale, light level, etc., also possible their combinations for range extension & call & -- incoming call acceptance \newline -- making a call on the intercom in the smart home\\
\hline

like, dislike & -- like / dislike or save/remove something (e.g. music track or video) \newline -- evaluating content to improve recommendations \newline -- expressing approval/disapproval for videoconference participants & mute & -- switching applications to silent mode \newline -- mute the microphone during a videoconference \newline -- switching the smart home to ``night mode"\\
\hline


stop, stop inverted & -- stop something (e.g. music or video playback) \newline -- swipe to scroll content when combined with the gesture ``stop inverted" & ok & -- command confirmation \newline -- confirmation of smart home mode switching\\
\hline

peace, peace inverted & -- switching the smart home to relax mode \newline -- activation something \newline -- rotation of the object when combined with the gesture ``peace inverted" & fist & -- expressing applause, approval or encouragement for videoconference participants \newline -- dragging objects when combined with the gesture ``palm" \\
\hline
two up, two up inverted & -- swipe to scroll content when combined with the gesture ``two up inverted" & rock & -- launching entertainment mode for smart home \newline -- changing the numerical value of any smart home/application characteristic to a maximum \\
\hline

\end{tabular}}
\end{center}
\caption{Possible uses of gestures. Gesture ``three2" has the same meaning as the gesture ``three" as the last is inconvenient to show by some people.}
\label{tabl:rules}
\end{table}

\begin{figure*}[htp]
  \centering
  \includegraphics[width=1.0\linewidth]{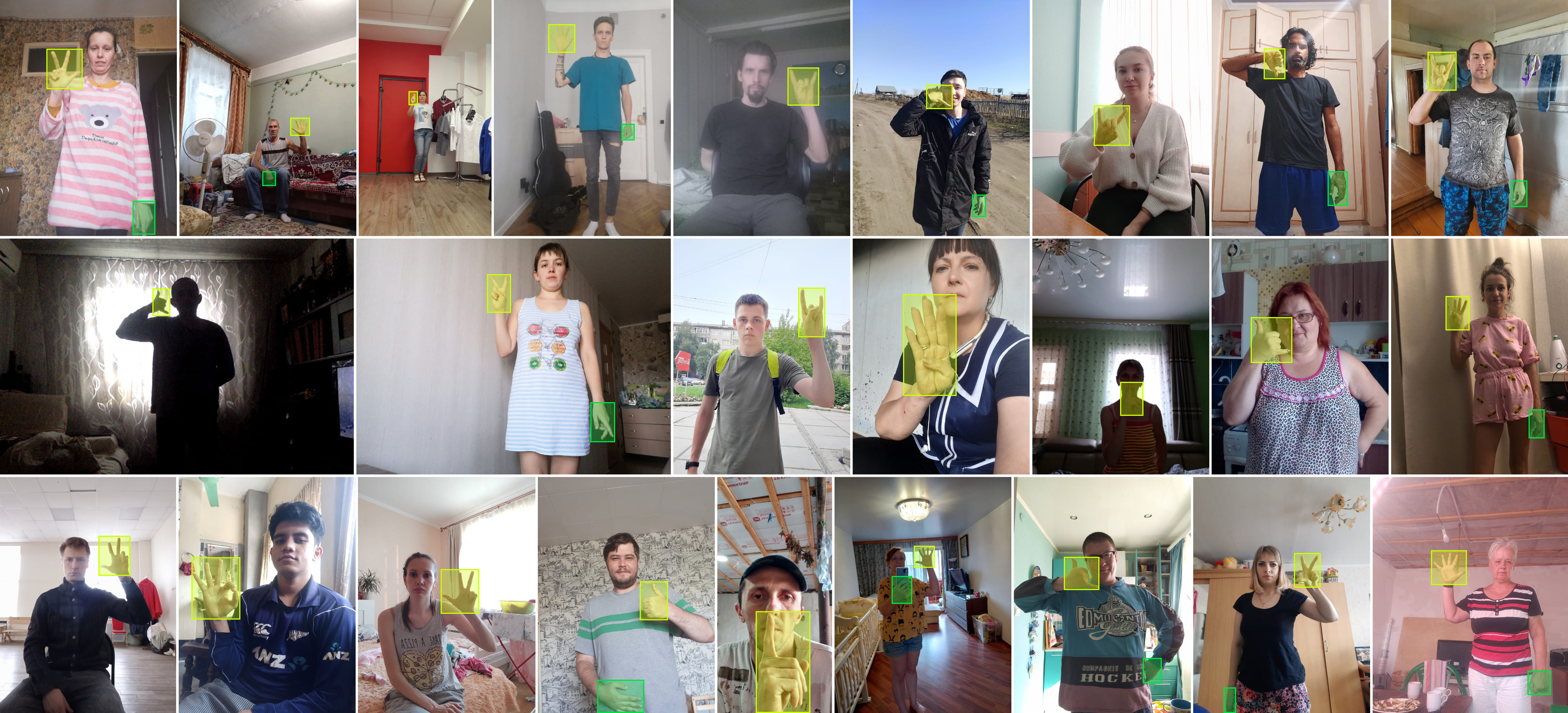}
  \caption{Examples of labeled samples from HaGRID. Gestures and ``no gestures" are highlighted in yellow and green bounding boxes, respectively.}
  \label{fig: samples}
\end{figure*}

\clearpage
\begin{table}[htp]
\centering
\begin{tabular}{lcccc}
\hline
Model & Weight Decay & Learning Rate & Scheduler & Scheduler' Params.\\
\hline
ResNet & $1^{-4}$ & $1^{-1}$ & ReduceLROnPlateau & mode: min, factor: 0.1\\
MobileNetV3  & $5^{-4}$ & $5^{-3}$ & StepLR & step size: 30, gamma: 0.1\\
VitB16 & $5^{-4}$ & $5^{-3}$ & CosineAnnealingLR & T max: 8\\
RetinaNet & $1^{-4}$ & $1^{-2}$ & StepLR & step size: 30, gamma: 0.1\\
SSDLite & $5^{-4}$ & $1^{-3}$ & StepLR & step size: 30, gamma: 0.1\\
YoloV7 & $5^{-4}$ & $1^{-2}$ & LambdaLR & sinusoidal function\\
\hline
\end{tabular}
\newline
\caption{Training hyperparameters.}
\label{tabl:params}
\end{table}
\end{document}